\documentclass{article}

\usepackage[preprint]{neurips_2024}

\usepackage[utf8]{inputenc} 
\usepackage[T1]{fontenc}    
\usepackage{hyperref}       
\usepackage{url}            
\usepackage{booktabs}       
\usepackage{amsfonts}       
\usepackage{nicefrac}       
\usepackage{microtype}      
\usepackage{xcolor}       
\usepackage{amsmath}
\usepackage{extarrows} 

\usepackage{tikz} 
\usetikzlibrary{shapes.geometric, arrows, calc, fadings, decorations.pathreplacing} 

\tikzset{%
  >=latex, 
  inner sep=0pt,%
  outer sep=2pt,%
  mark coordinate/.style={inner sep=0pt,outer sep=0pt,minimum size=3pt,
    fill=black,circle}%
}

\usepackage{textgreek} 
\usepackage{listings}  
\definecolor{dkgreen}{rgb}{0,0.6,0}
\definecolor{lviolet}{RGB}{231,154,255}
\definecolor{darkcyan}{RGB}{5, 55, 66}
\definecolor{dark_blue}{RGB}{3, 35, 42}
\title{Linearized Diffusion Map}

\author{%
  Julio Candañedo 
  \\
  \texttt{jcandane@asu.edu} \\
  \\
}

\newcommand{\R}{\mathbb{R}}
\DeclareUnicodeCharacter{2208}{\ensuremath{\in}}
\usepackage{newunicodechar}
\newunicodechar{≫}{\ensuremath{\gg}}
\newunicodechar{≪}{\ensuremath{\ll}}
\newunicodechar{∑}{\ensuremath{\sum}}
\newunicodechar{ℓ}{\ensuremath{\ell}}
\newunicodechar{∀}{\ensuremath{\forall}}
\newunicodechar{∈}{\ensuremath{\in}}
\newunicodechar{ℝ}{\ensuremath{\mathbb{R}}}
\newunicodechar{≥}{\ensuremath{\geq}}
\newunicodechar{≤}{\ensuremath{\leq}}
\newunicodechar{−}{\ensuremath{-}}
\newunicodechar{†}{{\textdagger}}
\newunicodechar{→}{\ensuremath{\rightarrow}}
\newunicodechar{×}{\ensuremath{\times}}
\newunicodechar{ε}{\ensuremath{\varepsilon}}
\newunicodechar{μ}{\ensuremath{\mu}}
\newunicodechar{∞}{\ensuremath{\infty}}

\begin{document}

\maketitle

\begin{abstract}
We introduce the Linearized Diffusion Map (LDM), a novel linear dimensionality reduction method constructed via a linear approximation of the diffusion-map kernel. LDM integrates the geometric intuition of diffusion-based nonlinear methods with the computational simplicity, efficiency, and interpretability inherent in linear embeddings such as PCA and classical MDS. Through comprehensive experiments on synthetic datasets (Swiss roll and hyperspheres) and real-world benchmarks (MNIST and COIL-20), we illustrate that LDM captures distinct geometric features of datasets compared to PCA, offering complementary advantages. Specifically, LDM embeddings outperform PCA in datasets exhibiting explicit manifold structures, particularly in high-dimensional regimes, whereas PCA remains preferable in scenarios dominated by variance or noise. Furthermore, the complete positivity of LDM's kernel matrix allows direct applicability of Non-negative Matrix Factorization (NMF), suggesting opportunities for interpretable latent-structure discovery. Our analysis positions LDM as a valuable new linear dimensionality reduction technique with promising theoretical and practical extensions.
\end{abstract}


\section{Introduction}

Dimensionality reduction is a foundational tool in data analysis, essential for simplifying, visualizing, and interpreting complex, high-dimensional datasets \cite{Hinton2006}. Given a dataset represented by a matrix \( R_{iX} \in \mathbb{R}^{N \times D} \), where indices \( i = 1,\dots,N \) label the samples and \( X = 1,\dots,D \) label the original (ambient) features, dimensionality reduction seeks a mapping:
\begin{equation}
    R_{iX} \;\rightarrow\; R_{ix},
\end{equation}
where \( R_{ix} \in \mathbb{R}^{N \times d} \), indexed by \( x = 1,\dots,d \), represents the embedded or latent representation, with \( d \ll D \). Since real-world datasets frequently exhibit \( D \gg d \), their intrinsic structures often remain opaque without effective dimensionality reduction techniques.

Historically, this concept was first formalized by Karl Pearson in 1901 through Principal Component Analysis (PCA) \cite{Pearson1901}, a linear approach that captures directions of maximum variance. PCA was complemented by classical Multidimensional Scaling, introduced by \cite{Torgerson1952}, explicitly focused on preserving global pairwise distances. However, both PCA and classical MDS struggle with datasets lying on intrinsically curved or nonlinear manifolds. This limitation motivated the development of nonlinear manifold-learning methods around the turn of the millennium, including Isomap \cite{isomap}. This algorithms explicitly recover nonlinear structures, effectively ``unrolling'' manifolds such as the iconic Swiss roll.

Further refinements emerged with the introduction of Diffusion Maps by Coifman and Lafon \cite{Coifman2006}, which provided a rigorous theoretical framework rooted in spectral theory and Markovian diffusion processes. Diffusion Maps define pairwise similarities through diffusion kernels and utilize eigen-decomposition, yielding embeddings that robustly preserve nonlinear geometric structures and mitigate noise sensitivity. Subsequently, t-distributed Stochastic Neighbor Embedding (t-SNE) \cite{Hinton2008} revolutionized exploratory visualization by prioritizing local structure preservation, facilitating intuitive cluster interpretation. More recently, Uniform Manifold Approximation and Projection (UMAP) \cite{McInnes2020} advanced the field by effectively balancing local and global structure preservation, combined with exceptional computational scalability.

Throughout this historical evolution—from PCA and classical MDS to manifold approaches such as Isomap, LLE, Diffusion Maps, t-SNE, and UMAP—dimensionality reduction techniques have consistently demonstrated broad utility, underpinning exploratory analysis, visualization, and preprocessing pipelines. By reducing complexity, these techniques improve computational efficiency, interpretability, and downstream task performance.

Despite significant advances in nonlinear dimensionality reduction, linear dimensionality reduction (LDR) methods retain considerable utility. Methods such as PCA and classical linear embeddings remain foundational due to their computational simplicity, interpretability, and robustness \cite{LDR}. Linear approaches typically involve fewer hyperparameters and yield stable embeddings, making them particularly valuable for reproducible statistical analysis and scientific research. Moreover, linear dimensionality reduction preserves explicit linear relationships essential for tasks such as regression analysis, anomaly detection, and efficient data compression.

Linear embeddings also provide substantial advantages as preprocessing steps for large-scale datasets. They facilitate rapid approximate nearest-neighbor (ANN) indexing, exemplified by indexing libraries such as FAISS \cite{FAISS}, enabling scalable downstream clustering and retrieval operations. Thus, while linear techniques may lag behind nonlinear methods in terms of direct clustering or manifold representation quality, their computational advantages and interpretability reinforce their indispensability in data analysis workflows.

Yet, although recent reviews provide extensive coverage of linear dimensionality reduction methods \cite{LDR}, a linear approximation or linearization of diffusion-map embeddings remains unexplored in the literature. This paper addresses precisely this gap by introducing and analyzing a Linearized Diffusion Map (LDM), a novel linear dimensionality reduction technique that combines aspects of diffusion geometry and linear embedding methods.

\subsubsection{Notation}

We summarize here the notation used throughout this paper. The kernel function is denoted as \( k(\cdot,\cdot) \), without indices. When evaluated over dataset points, it forms the \textit{unprocessed kernel-matrix}, denoted as \( k_{ij} = k(R_{iX}, R_{jX}) \). A \textit{processed} or \textit{proper} kernel-matrix, denoted \( K_{ij} \), refers to versions of \( k_{ij} \) that have been normalized or centered. The vector \(1_i\) represents an array of ones, and \(I_{ij}\) denotes the identity matrix.  
The dataset is described by an \textit{ambient} data matrix \(R_{iX}\), where samples are indexed by lowercase Latin indices \( i,j = 1,\dots,N \), and ambient-space features by uppercase indices \( X,Y = 1,\dots,D \). Correspondingly, \textit{latent embeddings} or reduced-dimension datasets are denoted by \( R_{ix} \), where latent-space features use lowercase indices \( x,y = 1,\dots,d \), with \( d \ll D \).
Matrix multiplication, including matrix-vector products, is denoted by the operator \( @ \), contracting the last index of the left-hand matrix with the first index of the right-hand matrix or vector, for example:
$
(K_{ij} @ R_{jx}) = \sum_j K_{ij} R_{jx}\quad
$.
Adjacent vectors or matrices with differing indices represent tensor (outer) products, e.g.,  $1_i 1_j = 1_i \otimes 1_j = 1_{ij}\,\,$,
while adjacent vectors or matrices sharing identical indices imply element-wise multiplication (Hadamard product), e.g.,
$1_i 1_i = 1_i \odot 1_i = 1_i\,\, $.

\section{PCA Theory}

Having established dimensionality reduction as a crucial analytical framework, we now turn to a detailed introduction to PCA, the foundational linear method from which much of dimensionality reduction has evolved. PCA’s enduring popularity lies in its elegant mathematical simplicity, interpretability, and computational efficiency, making it the standard reference point for both historical and contemporary techniques. Understanding PCA, and its nonlinear extension, Kernel PCA (kPCA), provides essential insight into how dimensionality reduction is practically implemented, laying a clear foundation for exploring more advanced and nonlinear methods in subsequent sections.

PCA begins with a dataset $R_{iX}\in \R^{N\times D}$, and centers the data by subtracting the mean feature vector $\mu_X = \frac{1}{N}\sum_{i=1}^NR_{iX}$, also known as the bias. That is our data is preprocessed to $C_{iX} = R_{iX} - 1_i \mu_X$. Next, we construct the covariance or linear kernel-matrix, this is linear and a quadratic form of this data:
\begin{align*}
    k_{ij} &= C_{iX}@C_{Xj} \\
    K_{ij} &= \frac{1}{N-1}\,C_{iX}@C_{Xj} \quad.
\end{align*}
Diagonalization of the proper kernel-matrix, on divided by Bessel's correction $\frac{1}{N-1}$, yields its own compression, in the form of an embedding, $R_{ix}$:
\begin{align*}
    K_{ij} R_{jx} &= \lambda_x R_{ix} \\
    K_{ij} &\approx \sum^d_x \lambda_x R_{ix} R^\dagger_{xj}\quad.
\end{align*}

\subsection{kPCA}

Next, a straight-forward generalization is to create the kernel-matrix from a nonlinear setting, this is achieved via a binary-function, two-input, kernel-function defined as $k(v_x, w_x)$  between two vectors, and it defines their similarity (instead of the linear cosine similarity).  
This function $k:X×X→\R$ is a valid kernel if and only if:
\begin{enumerate}
    \item Symmetry: $k(x,y)=k(y,x)$ for all $x,y∈X$,
    \item Positive Semi-Definiteness (PSD):    
    For any finite set ${x_1, x_2,…,x_n}$, the Gram matrix $k_{ij}=k(x_i,x_j)$ must satisfy:
    $∑_{i,j}^N = c_{i}c_{j} k_{ij} ≥ 0 \quad ∀c_{i},c_{j}∈\R$.
\end{enumerate}
Additionally some kernel-functions desire to have a probabilistic interpretation, for this we require element-wise positivity $k_{ij} \ge 0$.

Our focus would be on the Gaussian Radial-Basis-Function (RBF) for concreteness. We can construct the kernel-matrix with a collection of $N$ vectors, and their squared-Euclidean-distance-matrix between them $D^2_{ij} = R^2_{i}1_j + 1_i R^2_j - 2R_{iX}@R^\dagger_{Xj}$ (with $R^2_i = \sum_X R^2_{iX}$ and their decomposition by the law-of-cosines):
\begin{align}
k_{ij}(R_{iX}, R_{jX}) &= \exp{\left( - \frac{D^2_{ij}}{ε} \right)}_{ij} = \exp{\left( - \frac{|R_{iX} - R_{jX}|^2}{ε} \right)}_{ij} \\
&= \exp{\left( - \frac{R^2_{i}1_j + 1_i R^2_j - 2R_{iX}@R^\dagger_{Xj}}{ε} \right)}_{ij}\quad\quad.
\end{align}
The law-of-cosines decomposition of the distance-matrix is particularly useful, it relates an element-wise power (a nonlinear operation) to linear matrix operations!
Once we compute the kernel-matrix, $k_{ij}$ above, we need to double-center it to make it a proper kPCA kernel matrix $K_{ij}$:
\begin{align}
K_{ij} = k_{ij} - \frac{1_i}{N} \left( \sum_i k_{ij} \right)_j -  \left( \sum_j k_{ij} \right)_i \frac{1_j}{N} + \frac{1_i}{N} \left( \sum_{ij} k_{ij} \right) \frac{1_j}{N}\quad\quad.
\end{align}
The diagonalization of the kPCA kernel-matrix, similar to PCA, yields its own compression, and hence embedding:
\begin{align*}
    \label{kpca_centering}
    K_{ij}R_{jx} &= \lambda_x R_{ix} \\
    K_{ij} &\approx \sum^{d}_x \lambda_x R_{ix}R^\dagger_{xj}\quad.
\end{align*}

\subsection{Linearization of the RBF}

Although, the law-of-cosines was nice to factorize the distance-matrix it falls short of actually factorizing the element-wise application of the exponential function. In order to linearize this and hence factorize the kernel-matrix, we take it's the power series expansion:
\begin{align}
k_{ij} = \exp\left(-\frac{D^{2}_{ij}}{ε}\right) &= 1_{ij} - \frac{D^{2}_{ij}}{ε} + \frac{1}{2}\left(\frac{D^{2}_{ij}}{ε}\right)^2 - \cdots + \frac{(-1)^N}{N!} \left(\frac{D^{2}_{ij}}{ε}\right)^N  \\
&\approx 1_{ij} - \frac{D^{2}_{ij}}{ε} \\
k_{ij} &= 1_i 1_j - \frac{R^2_{i}1_j + 1_i R^2_j - 2R_{iX} @ R^\dagger_{Xj}}{ε}\quad.
\end{align}

This is approximately true for $ε > \max{D^2_{ij}}$, such that $k_{ij} \ge 0$ for all entries. Let $R^2_{\max} = \max{R^2_i}$, then $\max D_{ij}^2 ≤ (2R_{\max})^2 = 4R_{\max}^2$, therefore, for $ε = 4R_{\max}^2$ we have a successful linearization and hence factorization of the previously nonlinear kernel matrix. A notable point is that on large scales (or variance) any anisotropies in $ε$ disappear, and thus it is properly a scalar.

\subsection{Linearized kPCA}

Now, with our RBF-linearization we can apply this to our centralized kPCA kernel, eq. \ref{kpca_centering}.
Crucially, all constant terms cancel, terms with the dataset $R_{iX}$ remain and by making the tactical-substitution we obtain $N \mu_X = \sum_{k}R_{kX}$, we obtain:
\begin{align*}
K_{ij} &= \frac{2}{ε} \left( R_{iX} @ R^\dagger_{Xj} − 1_i μ_X @ R^\dagger_{Xj} − R_{iX} @ μ_{X} 1_j + 1_i 1_j μ_{X} @ μ_{x} \right) \\
&= \frac{2}{ε} \left( R_{iX} - 1_i μ_{X} \right)_{iX} @ \left( R^\dagger_{Xj} - μ_{X}1_{j} \right)_{Xj} \\
&= \frac{2}{ε} \,
C_{iX}C^\dagger_{Xj}
\end{align*}

Therefore, the Bessel correction used in PCA becomes $ε = 2(N−1)$, i.e. kPCA’s $ε$ absorbs this when $N$ is large. 

\subsection{MDS}

Now let's examine a Multi-Dimensional Scaling MDS. MDS applies a centering procedure (double centering) of the squared-Euclidean-distance-matrix, $D^2_{ij}$ as defined before.
With the resulting matrix $K_{ij}$, being symmetric and resembles a covariance matrix, and its diagonalization also yields the embedding coordinates. Let's substitute the law-of-cosines and simplify (Once again with tactical substitution $N μ_X = \sum_{k} R_{kX}$): 
\begin{align*}
K_{ij} &= -\frac{1}{2} H_{ik} D^{2}_{kl} H_{lj}, \quad \text{where} \quad H_{ij} = I_{ij} - \frac{1}{N}\, 1_{i}1_j^{†} \\
K_{ij} &= -\frac{1}{2}\, \left( I_{ik} - \frac{1}{N}\, 1_{i}1_k^{†}\right) \,@\, \left( R^2_{k}1_\ell + 1_k R^2_\ell - 2R_{kX} @ R^{†}_{X\ell} \right) \,@\, \left( I_{\ell j} - \frac{1}{N}\, 1_{\ell}1_j^{†} \right) \\
K_{ij} &= R_{iX} @ R^{†}_{Xj}−\frac{1}{N}\,∑_{k} R_{iX}R^{†}_{Xk} −\frac{1}{N}∑_{k} R_{kX}R_{Xj}^{†}+\frac{1}{N^2}∑_{kℓ} R_{kX}R_{Xℓ}^{†} \\
K_{ij} &= (R_{iX}−1_i μ_X) @ (R_{jX}−1_jμ_X)^{†}\quad.
\end{align*}
Amazingly, this is proportional to PCA and LkPCA with $ε=2$.
All the methods introduced here: PCA, MDS, or LkPCA, may be done efficiently, in $O(ND)$ time, if we use the LinearOperator, or the so-called matrix-free approach. Let's define the following LinearOperator, absorbing an arbitrary vector $v_j$ and yielding a new vector:
\begin{align}
    L^{\text{PCA}}_i(v_j) &= \frac{2}{ε} \, C_{iX} 
    \,@\,(C^\dagger_{Xj} 
    \,@\,v_j) \quad\quad.
\end{align}
We may obtain a diagonalization via a LinearOperator by using the Lanczos iterative algorithm, i.e. we begin with a guess vector, and with repeated application of this operator $L_i(v_j)$ we obtain the operator's spectrum.

\section{Linearized Diffusion Map}

\subsection{Diffusion Distances}

Thus far, PCA and MDS have been described as methods that operate primarily in Euclidean space. In contrast, Diffusion Maps introduce a fundamentally different metric space, the diffusion space, derived from a Markovian diffusion process. Given an unnormalized kernel-matrix $k_{ij}$, we construct a row-stochastic transition matrix:
\begin{align*}
    P_{ij} = \frac{k_{ij}}{\sum_j k_{ij}}\quad,
\end{align*}
interpreted as transition probabilities for a random walker moving across data points. This random-walk perspective naturally induces a new measure of distance, called the \emph{diffusion distance}, defined between points $i$ and $j$ at diffusion time $t$ as:
\begin{align*}
    D_t^2(i,j) = \sum_k \frac{\left(P_{ik}^t - P_{jk}^t\right)^2}{\pi_k}\quad,
\end{align*}
where $\pi_k$ is the stationary distribution of the Markov chain, satisfying $\sum_i \pi_i P_{ij} = \pi_j $. Intuitively, this metric measures similarity based on the probability distributions reached by the diffusion process after the $t$ steps, thus naturally capturing geometric structures on multiple scales.
Crucially, diffusion distances can be computed efficiently via spectral decomposition of the transition matrix:
\begin{align*}
    P_{ij}^t = \sum_{x=1}^{N} \lambda_x^t \psi_{ix}\phi_{jx}\quad,
\end{align*}
where $\psi_{ix}$ and $\phi_{jx}$ are the left and right eigenvectors, respectively, and $\lambda_x$ are eigenvalues ordered by magnitude ($ 1 = \lambda_1 \geq |\lambda_2| \geq \dots $). This spectral form immediately yields a low-dimensional embedding known as the \emph{Diffusion Map embedding}, defined as:
\begin{align*}
    R_{ix}^{\text{DM}}(t) = \lambda_x^t \psi_{ix}\quad,
\end{align*}
ensuring that the Euclidean distances in the embedded space approximate the original diffusion distances at the diffusion time $t$.
The diffusion process itself can be intuitively understood as analogous to simulated annealing from statistical mechanics. At small diffusion times ($t \rightarrow 0 $), the distances emphasize fine-grained local geometric details. As diffusion progresses ($t$ increases), local irregularities are smoothed out, revealing stable global structures. Ultimately, as $ t \rightarrow \infty $, the diffusion process converges to the stationary distribution, effectively "cooling" to equilibrium and losing all memory of the initial structures.

The transition matrix can be constructed from the RBF kernel in two distinct normalization schemes, with $k_i = \sum_j k_{ij}$:
\begin{align*}
    K_{ij} &= k_i^{-1/2} k_{ij} k_j^{-1/2} \quad \text{ symmetric,} \\
    P_{ij} &= k_i^{-1} K_{ij}, \quad\quad\quad \text{ asymmetric,}.
\end{align*}
The symmetric normalization facilitates convenient spectral decompositions, allowing diffusion-maps to effectively characterize manifold geometries in the data. However, both could be linearized and used as a LDR method.

\subsection{Linearized Diffusion Map}

We may begin with the RBF-kernel-matrix and its subsequent linearization. With this we are left with the following kernel operator:
\begin{align*}
k_{ij}
&= \left( 1_i - \frac{R^2_i}{ε}  \right) 1_j - 1_i \frac{R^2_j}{ε} + \frac{2}{ε} \, R_{iX} @ R^{\dagger}_{Xj} \quad.
\end{align*}
The diffusion-map normalization is computed with the following linear operator, $k_i = L_i(1_j)$:
\begin{align}
    \mathcal{L}_i(v_j) &= k_{ij}v_j = \left( 1_i - \frac{R^2_i}{ε}  \right) 1_j v_j - 1_i \frac{R^2_j v_j}{ε} + \frac{2}{ε} \, R_{iX} @ ( R^{\dagger}_{Xj} @ v_j )\quad. \\
    N_i &= \left( L_i(1_j) \right)^{-1/2} \quad\quad\text{symmetric,} \\
    N_i &= \left( L_i(1_j) \right)^{-1} \quad\quad\text{  asymmetric.}
\end{align}
Once the normalization is obtained this can be applied in two different ways, obtaining two kinds of Linearized-Diffusion-Map (LDM) models. We call the symmetric-diffusion-map as the default and call this the LDM, while the asymmetric version is LDM-Asymmetric (LDM-A):
\begin{align}
L^{\text{LDM}}_i(v_j) &= N_i\left( 1_i - \frac{R^2_i}{ε}  \right) N_j @ v_j - N_i \frac{R^2_j @ (N_j v_j)}{ε} + \frac{2}{ε} \, N_i R_{iX} @ ( R^{\dagger}_{Xj} @ (N_j v_j )) \\
L^{\text{LDM-A}}_i(v_j) &= N_i\left( 1_i - \frac{R^2_i}{ε}  \right) 1_j @ v_j - N_i \frac{R^2_j @ v_j}{ε} + \frac{2}{ε} \,N_i R_{iX} @ ( R^{\dagger}_{Xj} @ v_j )
\end{align}
For both theories, the parameters are: $\{ N_i, 1_i, R^2_i, R_{iX}, ε \}$
Just like PCA, full diffusion-map, LDM uses these linear-operators to facilitate efficient diagonalization to obtain the LDM embedding.


\section{Datasets and Experimentation}

\subsection{The Swiss Roll}

Let us do a visual comparison; for this we use the infamous Swiss-roll dataset, and compare the LDR techniques: PCA and our method LDM. Both linear techniques cannot unroll the Swiss-roll as expected, but they yield different results. While PCA projects a spiral curve, LDM has made some effort to unroll it, albeit incompletely, we have folded sheets, instead of one sheet. Hence, LDM can provide different projections than PCA, for clustering or interpretations. 

\begin{figure}[!hbt]
    \centering
    \includegraphics[width=\linewidth]{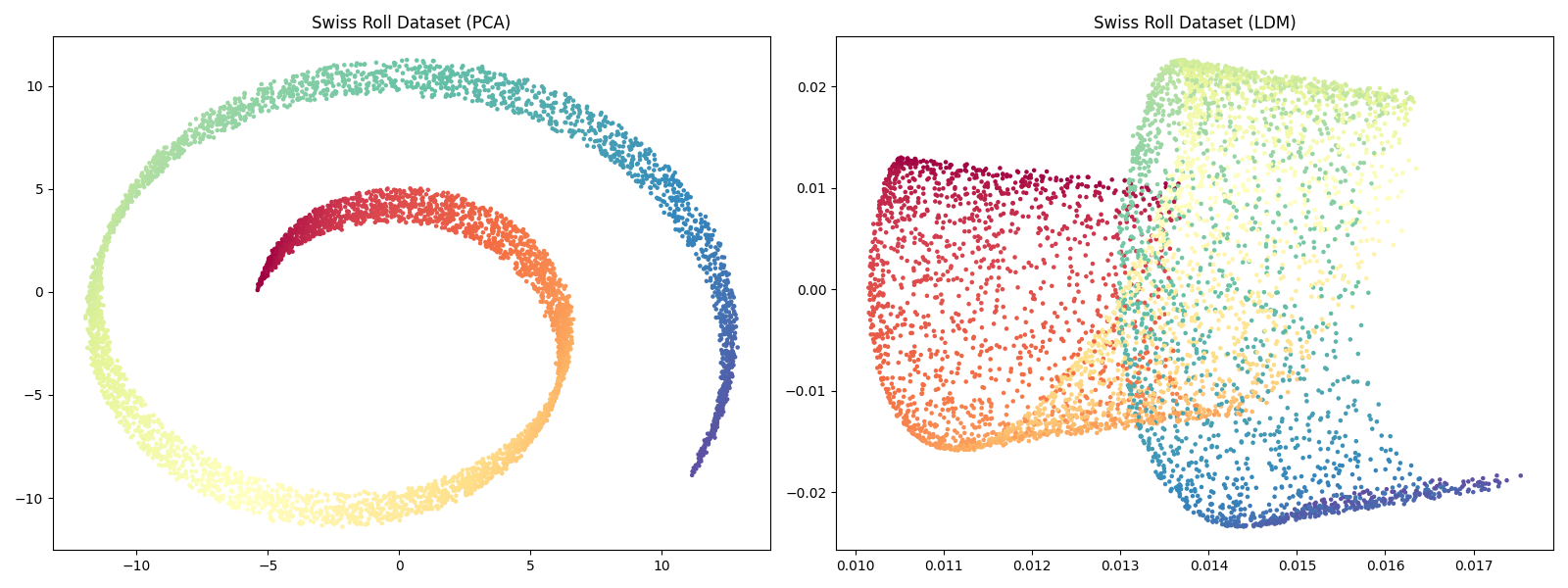}
    \caption{A visual comparison between embeddings of: PCA (left plot) and LDM (right plot). Colors indicate the original proximity of points before projection.}
    \label{fig:Swiss}
\end{figure}

\subsection{MNIST and COIL-20 Datasets}

To extend our comparisons beyond synthetic datasets, we evaluate PCA and LDM performance on two well-known real-world benchmark datasets: MNIST and COIL-20. 
The MNIST dataset consists of 60,000 grayscale images of handwritten digits (0--9), each image having dimensions $28 \times 28$ pixels, resulting in a dataset with ambient dimensionality $D = 784$. COIL-20 (Columbia Object Image Library) is a dataset containing grayscale images of 20 distinct physical objects. Each object was photographed at 72 equally-spaced angles (5-degree increments), resulting in $20 \times 72 = 1440$ images. Each image has a resolution of $128 \times 128$ pixels, yielding an ambient dimensionality of $D = 16,384$ per image.

Our goal is to efficiently compute approximate nearest-neighbor (ANN) lists using the unsupervised linear embedding methods PCA and LDM. Given a dataset with $N$ points, computing exact $k$-nearest neighbors (kNN) involves constructing a neighbor list matrix $N_{ik}\in\mathbb{N}^{N\times k}$, where each row $i$ contains the indices of the $k$ nearest neighbors of point $i$. Exact computation of these neighbor lists typically has complexity $O(N^2D)$, which quickly becomes prohibitively expensive for large-scale data. Thus, linear-complexity embedding techniques (such as PCA or LDM), with computational cost $O(ND)$, are valuable tools for approximate nearest-neighbor retrieval tasks.

\begin{figure}[ht]
    \centering
    \includegraphics[width=0.49\linewidth]{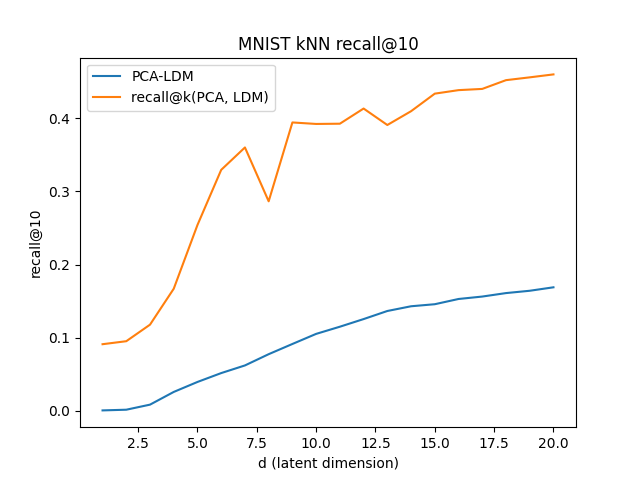}
    \includegraphics[width=0.49\linewidth]{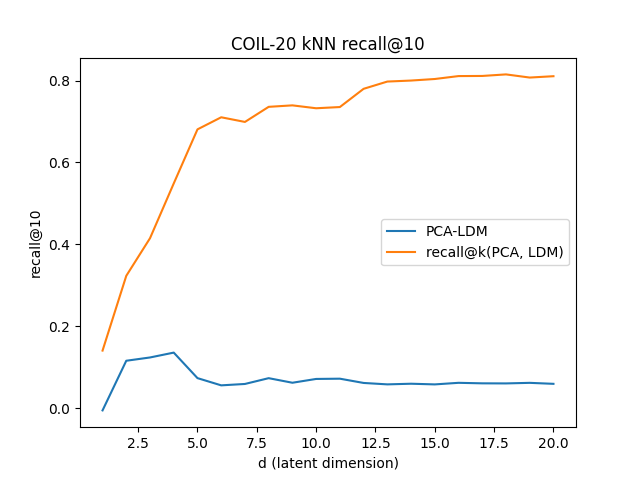}
    \caption{
    Comparison of PCA and LDM embeddings for approximate nearest-neighbor retrieval tasks on MNIST (\textbf{left}) and COIL-20 (\textbf{right}). Each plot displays the difference in recall@10 scores (PCA minus LDM) relative to exact ground-truth kNN neighbors, as a function of latent dimension $d$. Also shown (dashed curve) is the recall@10 overlap directly comparing PCA and LDM neighbor lists, indicating the extent of their common retrieved neighbors across varying embedding dimensions.
    }
    \label{fig:datasets}
\end{figure}

To assess embedding quality, we measure the agreement between approximate neighbor lists generated by PCA ($N^\text{PCA}_{ik}$) or LDM ($N^\text{LDM}_{ik}$) and the exact ground-truth neighbors ($N^\text{kNN}_{ik}$). We quantify neighbor-list accuracy using the \textit{recall@k} metric, defined as:
\begin{align*}
\text{recall@k}(N_{ik}, N'_{ik}) = \frac{1}{N}\sum_{i=1}^{N}\frac{|N_{ik}\cap N'_{ik}|}{k},
\end{align*}
which measures the average fraction of neighbors correctly retrieved by the approximate method relative to the ground-truth neighbors.
Since $N$ and $D$ are fixed for these datasets, we analyze how varying the latent dimension $d$ affects the neighbor-list recall for both PCA and LDM embeddings. The results are presented in Figure~\ref{fig:datasets}, which clearly illustrates the relative strengths and weaknesses of LDM versus PCA embeddings on real-world data.

\subsection{Random Hypersphere Sampling}

The hypersphere dataset provides an ideal testing ground for comparing dimensionality reduction methods because its intrinsic geometry is explicit, well-understood, and parametrically controllable. Specifically, we sample $N$ points uniformly at random from the surface of a high-dimensional sphere embedded in ambient dimension $D$, optionally adding small radial noise. Formally, the dataset is constructed as follows:
\[
X_i \sim \mathcal{N}(0, I_D), \quad X_i \leftarrow \frac{X_i}{\|X_i\|}, \quad i=1,\dots,N \quad.
\]
This procedure generates points uniformly distributed on a $(D-1)$-dimensional spherical manifold embedded within $\mathbb{R}^{D}$. Crucially, this dataset possesses a clearly defined intrinsic dimensionality ($d = D-1$), allowing us to fix a target latent dimension and systematically vary both the number of samples $N$ and the ambient dimension $D$. By comparing embeddings obtained from PCA and Linearized Diffusion Maps (LDM) on this hypersphere dataset, we explicitly test how effectively each method captures intrinsic nonlinear manifold geometry. This experiment highlights the strengths and limitations of LDM relative to PCA, particularly illuminating scenarios where linearized diffusion geometry offers meaningful improvements over classical linear variance maximization.

\begin{figure}[ht]
    \centering
    \includegraphics[width=\linewidth]{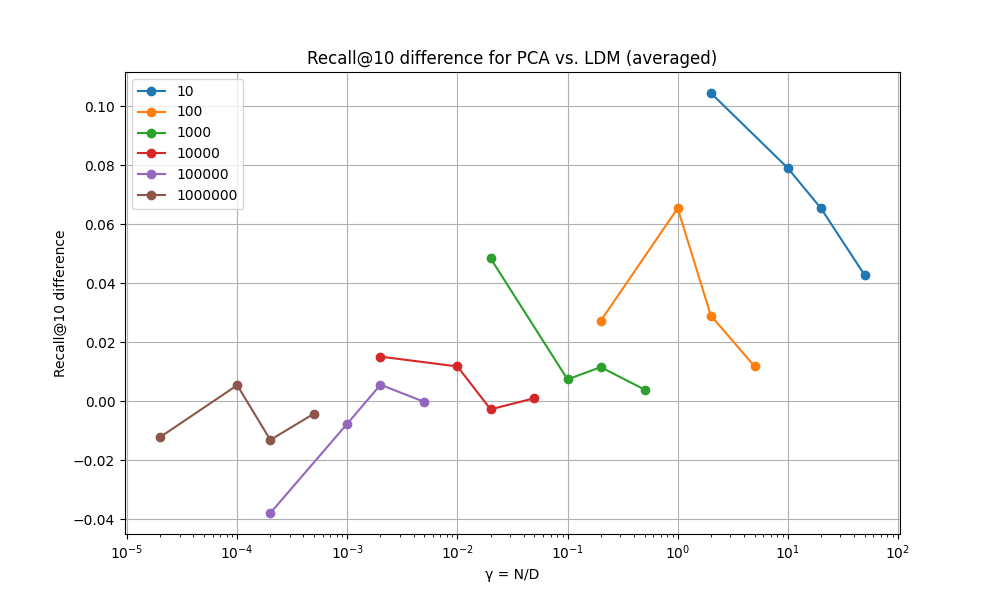}
    \includegraphics[width=\linewidth]{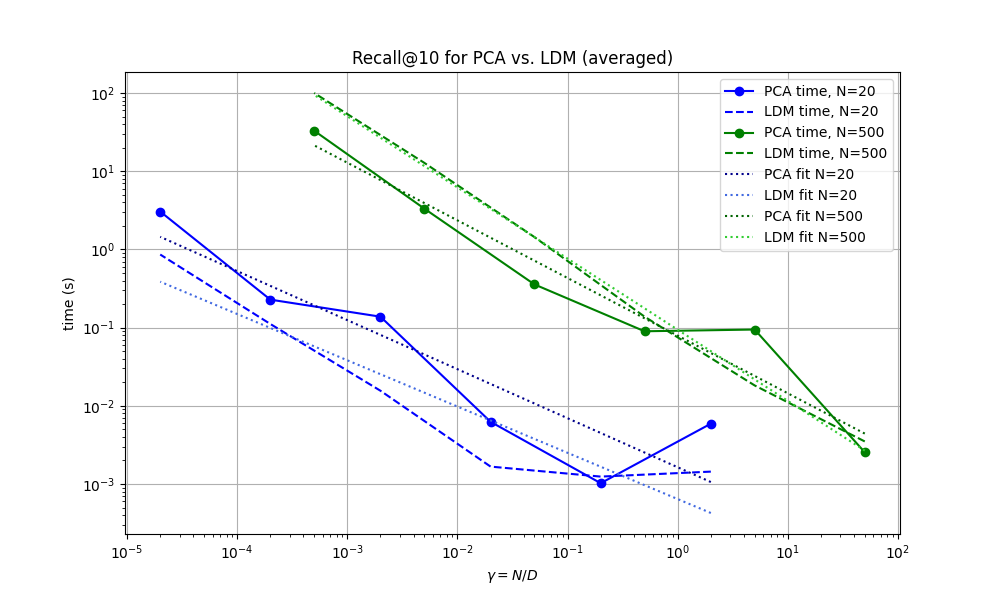}
    \caption{Comparison of PCA and LDM performance on hypersphere datasets. \textbf{Top:} Difference in recall@10 scores (PCA minus LDM) relative to the ground-truth kNN, plotted as a function of the ratio $\gamma = N/D$. Positive values indicate PCA outperforming LDM, while negative values indicate LDM superiority. \textbf{Bottom:} Computational time complexity (in seconds) for PCA and LDM embeddings, showing linear scaling behavior with respect to dataset size.}
    \label{fig:hypersphere}
\end{figure}

The results of this comparison are shown in Figure~\ref{fig:hypersphere}. The upper plot illustrates the difference in ANN neighbor-list quality between PCA and LDM compared to the ground-truth kNN neighbors, specifically measuring:
\[
\text{recall@10}(N^\text{PCA}_{ik}, N^\text{kNN}_{ik}) - \text{recall@10}(N^\text{LDM}_{ik}, N^\text{kNN}_{ik}),
\]
as a function of dataset size ratio $\gamma = \frac{N}{D}$. These results demonstrate that LDM increasingly outperforms PCA as the ambient dimensionality $D$ grows large, while PCA generally remains superior for generating ANN lists in regimes with larger sample sizes $N$ relative to the ambient dimension $D$. The lower plot confirms the linear computational complexity of both PCA and LDM with respect to the dataset size, reinforcing the computational feasibility of both techniques.

\section{Conclusion}

In this paper, we introduced the Linearized Diffusion Map (LDM), a novel linear dimensionality reduction method derived from a linear approximation of the diffusion-map kernel. LDM leverages diffusion geometry's local manifold-preserving characteristics while maintaining the computational efficiency and interpretability associated with linear embedding methods such as PCA and MDS. Through comprehensive experimental comparisons—including synthetic datasets (Swiss roll and hypersphere) and real-world benchmarks (MNIST and COIL-20)—we demonstrated that LDM provides meaningful geometric embeddings distinct from PCA.

Our experiments indicate that LDM excels particularly in scenarios with explicit global manifold structure, such as high-dimensional spheres or curved synthetic manifolds, clearly outperforming PCA in terms of nearest-neighbor retrieval accuracy (recall@k). However, on realistic, noisy, or less explicitly structured datasets like MNIST and COIL-20, PCA remains generally superior for generating approximate nearest-neighbor lists. These observations suggest that LDM and PCA can be viewed as complementary tools rather than strict competitors. Indeed, LDM could potentially be used alongside PCA as a secondary embedding strategy, enhancing ANN discovery by capturing different aspects of the underlying data geometry.

Building upon these insights, several promising directions emerge for future exploration. First, the diffusion-map kernel scale parameter $\varepsilon$, initially treated as a global constant, could be adaptively or locally defined as a point-dependent scale $\varepsilon_i$. Such an adaptive kernel would enable LDM to account more effectively for local geometric heterogeneity, potentially enhancing its performance on datasets exhibiting varying local densities or manifold curvatures. The necessary pointwise normalization factors computed during the diffusion-map construction could provide a principled approach for determining these local scales in a data-driven manner.

Moreover, the complementary strengths of PCA and LDM suggest integrating these linear embedding methods within hybrid workflows to leverage their distinct geometric sensitivities. Specifically, PCA embeddings, optimized for capturing variance-dominated directions, could serve as an initial step, quickly narrowing candidate neighborhoods. Subsequently, LDM embeddings—capturing nuanced diffusion-geometry-informed relationships—could refine these neighborhoods, improving approximate nearest-neighbor (ANN) retrieval tasks through this combined hierarchical embedding approach.

Lastly, the inherent positivity and linear factorization of the LDM kernel matrix offer intriguing theoretical and practical implications. Because the linearized kernel is completely positive, it becomes directly amenable to Non-negative Matrix Factorization (NMF). This opens up new possibilities for interpretable latent structure discovery, enabling applications in domains where interpretability and explicit factorization of latent features are crucial, such as bioinformatics, recommender systems, or topic modeling. Future work could explore precisely how LDM-derived kernels might improve the quality or interpretability of NMF decompositions, potentially uncovering novel latent structures otherwise obscured by traditional linear or nonlinear dimensionality-reduction approaches.

\bibliographystyle{fancybib}
\bibliography{refs}

\end{document}